\pdfoutput=1

\documentclass[11pt]{article}
\usepackage{acl}

\usepackage{times}
\usepackage{latexsym}

\usepackage{microtype}

\usepackage{url}            
\usepackage{booktabs}       
\usepackage{amsmath}
\usepackage{amsfonts}       
\usepackage{nicefrac}       
\usepackage{natbib}
\usepackage{array}
\usepackage{makecell}
\usepackage{multirow}
\usepackage{amsfonts}
\usepackage{graphicx}
\usepackage{float}
\usepackage{caption} 
\usepackage{mathtools}
\usepackage[autostyle]{csquotes}
\usepackage{caption}
\usepackage{subcaption}
\usepackage{wrapfig}

\usepackage{tablefootnote}

\makeatletter
\newcommand{\thickhline}{%
    \noalign {\ifnum 0=`}\fi \hrule height 1pt
    \futurelet \reserved@a \@xhline
}

\newcommand{\NoGap}{\vspace{-5mm}}  
\newcommand{\ours}{TrSum}
\newcommand{\extref}{ExtRef}

\title{Transductive Learning for Abstractive News Summarization}
\date{}

\author{Arthur Bra\v{z}inskas$^1$\, Mengwen Liu$^2$ \, Ramesh Nallapati$^3$ \, Sujith Ravi$^2$\, Markus Dreyer$^2$ \bigskip \\ 
 $^1$ILCC, University of Edinburgh \, $^2$Amazon Alexa \, $^3$AWS AI Labs \\
\texttt{abrazinskas@ed.ac.uk} \\ \texttt{\{mengwliu, rnallapa, sujithai, mddreyer\}@amazon.com}}

\begin{document}

\setlength{\belowdisplayskip}{5pt}
\setlength{\abovedisplayskip}{5pt}

\maketitle

\begin{abstract}

Pre-trained and fine-tuned news summarizers are expected to generalize to news articles unseen in the fine-tuning (training) phase. However, these articles often contain specifics, such as new events and people, a summarizer could not learn about in training. This applies to scenarios such as a news publisher training a summarizer on dated news and summarizing incoming recent news. In this work, we explore the first application of \textit{transductive learning} to summarization where we further fine-tune models on test set inputs. Specifically, we construct pseudo summaries from salient article sentences and input randomly masked articles. Moreover, this approach is also beneficial in the fine-tuning phase, where we jointly predict extractive pseudo references and abstractive gold summaries in the training set. We show that our approach yields state-of-the-art results on CNN/DM and NYT datasets, improving ROUGE-L by 1.05 and 0.74, respectively. Importantly, our approach does not require any changes of the original architecture. Moreover, we show the benefits of transduction from dated to more recent CNN news. Finally, through human and automatic evaluation, we demonstrate improvements in summary abstractiveness and coherence. 

\end{abstract}

\section{Introduction}
\label{seq:intro}

Language model pre-training has advanced the state-of-the-art in many NLP tasks ranging from sentiment analysis, question answering, natural language inference, named entity recognition, and textual similarity; more recently, they have been used in summarization \citep{liu2019text, lewis2019bart}. State-of-the-art pre-trained models include \textsc{GPT} \citep{radford2018improving}, \textsc{BERT} \citep{devlin2018bert}, \textsc{BART} \citep{lewis2019bart}, \textsc{Pegasus} \citep{zhang2019pegasus}. 


\begin{table}[t!]
    \centering
 	\footnotesize 
    \begin{tabular}{ >{\centering\arraybackslash} m{1.5cm} m{5cm}} 
 \thickhline
    \textbf{Abstractive} & \vspace{0.5em}  
     The penalty is more than 10 times the previous record, according to a newspaper report. Utility commission to force Pacific Gas \& Electric Co. to make infrastructure improvements. Company apologizes for explosion that killed 8, says it is using lessons learned to improve safety. \vspace{0.5em}  \\ \thickhline
    \textbf{Extractive} & \vspace{0.5em} 
     The California Public Utilities Commission on Thursday said it is ordering Pacific Gas \& Electric Co. to pay a record \textcolor{red}{\$1.6 billion penalty for unsafe operation} of its gas transmission system, including the \textcolor{orange}{pipeline rupture that killed eight people in San Bruno} in September 2010. Most of the penalty amounts to \textcolor{blue}{forced spending on improving pipeline safety}. On September 9, 2010, a section of PG\&E pipeline exploded in San Bruno, killing eight people and injuring more than 50 others.
    \vspace{0.5em} \\ \thickhline
\textbf{Ours} & \vspace{0.5em} 
     Pacific Gas \& Electric Co. is ordered to pay a record \textcolor{red}{\$1.6 billion penalty}. Most of the penalty amounts to \textcolor{blue}{forced spending on improving pipeline safety}. A section of PG\&E pipeline \textcolor{orange}{exploded in San Bruno in 2010}, \textcolor{orange}{killing eight people}. The company says it is working to become the safest energy company in the U.S.
    \vspace{0.5em} \\ \thickhline
    \end{tabular}
    \caption{Example summaries that are human-written (abstractive), and produced by extractive and our systems. Colored text indicates important details not present in the human-written summary.}
    \label{table:front-example-summ}
\NoGap
\end{table}

%
%

These models acquire prior syntactic and semantic knowledge from large text corpora and are further fine-tuned on task-specific smaller datasets, such as news article-summary pairs. However, specifics of test set news articles might not be well represented in the training set. For example, a news publisher might train a summarizer on dated news and wants to summarize latest incoming news. This suggests potential improvements if the summarizer learns these specifics before summaries are generated. In this work, we explore \textit{transductive learning}~\citep{vapnik1998statistical} by adapting a fine-tuned summarizer to the test set by learning from its input articles.

The main obstacle for \textit{transduction} is the absence of a reliable training signal, as no references are available in test time. Therefore, we propose constructing extractive pseudo references by selecting salient sentences from the input text with a separately trained model. Salient sentences are often fused and compressed to form abstractive summaries~\citep{lebanoff2019scoring}, and contain additional important details providing better context, as illustrated in Table~\ref{table:front-example-summ}. Further, we use a denoising objective to predict salient sentences conditioned on masked input articles. In this way, the model balances the copying and generation dynamic~\cite{see2017get, gehrmann-etal-2018-bottom, brazinskas2020-unsupervised} as not all information for accurate summary predictions is available in the masked input. To further preserve summary abstractivness, we predict a small portion of abstractive summaries ($\sim$5\% on CNN/DM) from the annotated training set. This results in only a small fraction of the training time needed to perform transduction ($<$ 4\% on CNN/DM\footnote{On an AWS 8-GPU p3.8xlarge instance, full training took 9 hours while transduction only 15 minutes.}). Moreover, we leverage salient sentences from training set inputs in the fine-tuning phase by predicting both abstractive and extractive references. As we show, this method outperforms standard fine-tuning on abstractive references alone. Finally, we show improvements in the scenario when only dated news articles with summaries are available for training and the aim is to summarize recent news articles in test time.



All in all, we empirically demonstrate that our model (\textsc{\ours{}}), that utilizes salient sentences in the fine-tuning and transduction phases, significantly improves the quality of summaries. Besides achieving state-of-the-art results on standard datasets (CNN/DM~\citep{hermann2015teaching} and NYT~\citep{sandhaus2008new}), it also yields more coherent and abstractive summaries. Our main contributions can be summarized as follows.


\begin{itemize}
    \item we present the first application of transductive learning to summarization;
    \item we show state-of-the-art results on standard summarization datasets (CNN/DM and NYT);
    \item we show that transduction is beneficial for summarizing more recent CNN news.\footnote{The codebase will be publicly available.}
\end{itemize}



\begin{figure}
    \centering
    \includegraphics[width=0.45\textwidth]{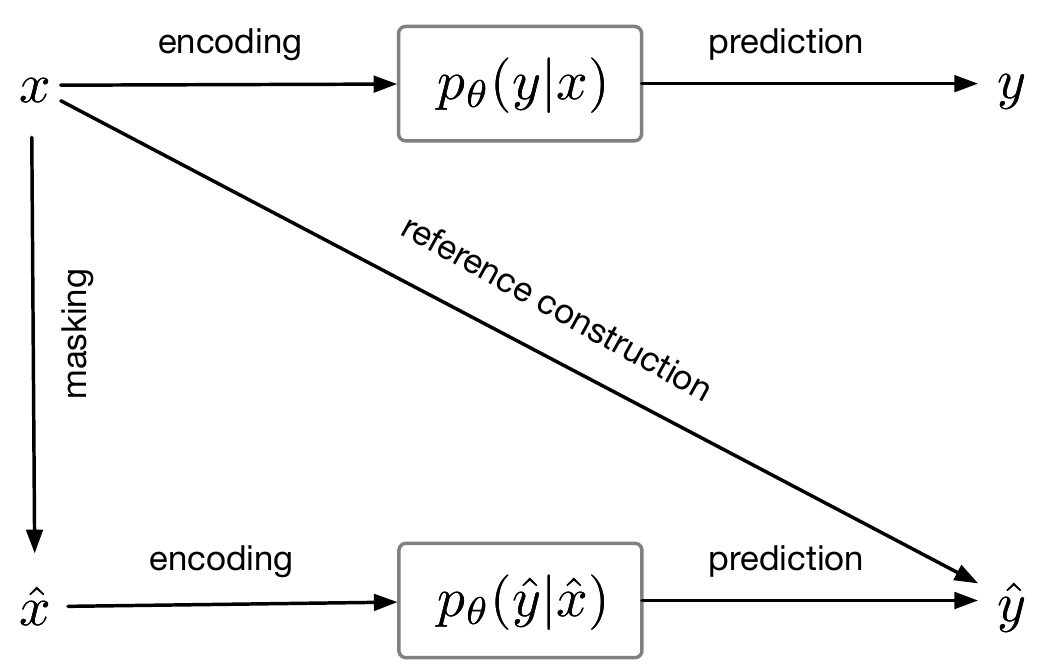}
    \caption{Illustration of the joint objective and the associated procedure. Here we randomly mask the input article $x$ resulting in $\hat x$. Further, construct $\hat y$ by concatenating salient sentences in $x$. Lastly, we jointly predict abstractive and extractive references $y$ and $\hat y$, respectively.}
    \label{fig:joint_objective}
    \NoGap{}
\end{figure}

\section{Joint Fine-Tuning}
\label{sec:joint_fine_tuning}

Our model (\textsc{\ours{}}) has a Transformer encoder-decoder architecture~\citep{vaswani2017attention}, which is initialized with pre-trained BART \citep{lewis2019bart}. Before we learn from the test set articles using transductive learning (presented in Sec.~\ref{sec:transduction}), we jointly fine-tune the model on extractive and abstractive references in the training set. Extractive references are useful for learning, as they often contain omitted details in abstractive summaries and provide additional context to the reader, see Table~\ref{table:front-example-summ}.


Let $\{x_i, y_i\}_{i=1}^N$ be article-summary pairs in the training set. First, we greedily select $k$ sentences from the input article $x$ that maximize the ROUGE score\footnote{We used the average of ROUGE-1 and ROUGE-2 F scores.} to the summary $y$ by following \citet{liu2019text}. We concatenate these sentences to form an extractive summary $\hat{y}$ that is word-by-word predicted using teacher-forcing~\citep{williams1989learning}. Further, to prevent trivial solutions, we randomly mask words in $x$ with a special mask token.\footnote{We also experimented with masking only salient sentences. However, this lead to inferior results.} Intuitively, this forces the decoder to balance between copying from the input and generating novel content~\citep{see2017get, gehrmann-etal-2018-bottom, brazinskas2020-unsupervised}. This procedure creates pairs $\{\hat x_j, \hat y_j\}_{j=1}^M$, and we formulate the \textit{joint fine-tuning} objective in Eq.~\ref{eq:joint-fine-tuning}. We also illustrate the whole procedure in Fig.~\ref{fig:joint_objective}.
\begin{equation}
\begin{aligned}
     \dfrac{1}{N}\sum_{i=1}^{N} \log p_{\theta}(y_i|x_i) + \dfrac{1}{M}\sum_{j=1}^{M} \log p_{\theta}(\hat{y}_j|\hat{x}_j)
\end{aligned}
\label{eq:joint-fine-tuning}
\end{equation}

Notice that the joint objective in Eq.~\ref{eq:joint-fine-tuning} re-uses the model's architecture without a specialized task embedding. The model can easily differentiate between abstractive and extractive summary prediction/generation as only in the latter the input contains a special mask token. We validate this in an ablation experiment presented in Sec.~\ref{sec:ablation}. 

Lastly, our main goal is to learn an abstractive summarizer $p_{\theta}(y|x)$ without overfitting on extractive references. Thus, we control for the ratio of abstractive and extractive instances $N$ and $M$, respectively, by drawing decisions from the Bernoulli distribution $Bern(\alpha)$. If $\alpha$ is set to 0, it results in abstractive pairs only.

\section{Transduction}
\label{sec:transduction}

Consider a scenario where a news publishing agency has a fine-tuned model on dated article-summary pairs and wants to summarize upcoming news articles for which summaries are not yet available. In this setting, an immediate response might not be necessary and latency can be traded for summary quality. In this light, we propose to leverage \textit{transductive learning}~\citep{vapnik1998statistical} and further fine-tune the model by learning from test set input articles. First, we train an extractive summarizer that predicts salient sentences, as explained in Sec.~\ref{sec:extr-summ}. Second, we extract salient sentences from test set input articles and construct pseudo references $\hat y$. Lastly, we optimize the model by predicting these references using $p_{\theta}(\hat y | \hat x)$ in Eq.~\ref{eq:joint-fine-tuning}.

\subsection{Extractive Summarizer}
\label{sec:extr-summ}

\begin{figure}
    \centering
    \includegraphics[width=0.4\textwidth]{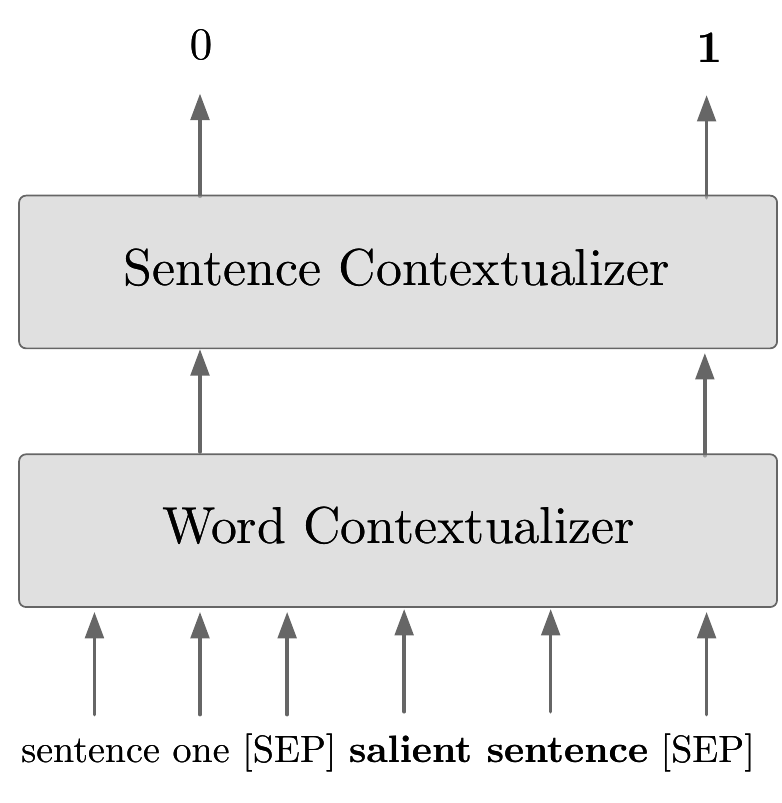}
    \caption{Extractive summarizer contextualizes words and subsequently sentences. The final outputs are binary tags where 1 indicates a salient sentence.}
    \label{fig:tagger}
    \NoGap{}
\end{figure}

To produce extractive references on the test set, we train an extractive summarizer. The summarizer consists of two Transformer encoders and predicts which sentences are salient, as illustrated in Fig.~\ref{fig:tagger}. Formally, let [$s_1$, $s_2$, ..., $s_m$] denote sentences in an article where each sentence is separated by a special symbol (\textsc{[SEP]}). Further, let [$b_1$, $b_2$, ..., $b_m$] be their associated binary tags where $1$ indicates a salient sentence. 

To compute model predictions for sentences, we proceed as follows. First, we feed the sequence of concatenated sentences [$s_1$, $s_2$, ..., $s_m$] to the first encoder and obtain sentence representations [$e_1$, $e_2$, ..., $e_m$]. Intuitively, these representations capture semantics of each sentence useful for determining their salience and how well they summarize the whole article. To better capture cross-sentence dependencies, we feed the sentence representations to the second encoder and obtain contextualized representations [$c_1$, $c_2$, ..., $c_m$]. Finally, we feed each representation $c_i$ to a feed-forward neural network $f_{\theta}(c_i)$ to obtain scores.

\subsection{Regularization}
\label{sec:trans-reg}
In transduction, when the model is solely optimized for predicting extractive summaries, the previously learned abstractive summarization and its performance can degradate~\citep{goodfellow2013empirical, kemker2017measuring}; see Sec.~\ref{sec:ablation} for a confirming experiment. As a form of regularization, we propose to additionally predict abstractive summaries from the training set using the joint objective in Eq.~\ref{eq:joint-fine-tuning}. In practice, we found that sampling a similar amount of training pairs as in the test set (about 5\% on CNN/DM) to be sufficient. This regularization is conceptually similar to \textsc{RecSum}~\citep{chen-etal-2020-recall} where the model is penalized for parameters deviating from the original ones. Another point of consideration is that the extractive summarizer, presented in Sec.~\ref{sec:extr-summ}, can erroneously select non-salient sentences, resulting in less reliable references. Consequently, we found it beneficial to also add extractive pairs from the training set created using a heuristic presented in Sec.~\ref{sec:joint_fine_tuning}.

\subsection{Tracking of Overfitting}
\label{sec:trans-val}
Tracking of overfitting is essential for model development. To monitor overfitting during transduction, we propose the following simple procedure. First, we sample a tiny subset of validation pairs (around 1,000). To closely resemble transduction, we produce extractive references using the extractive model presented in Sec.~\ref{sec:extr-summ}. Further, we combine the validation extractive pairs with the training and test set pairs used for transduction (see Sec.~\ref{sec:trans-reg}). Finally, we track ROUGE-L scores on the validation human-written abstractive references. This, in turn, allows us to determine when abstractive summarization performance starts to decrease to perform early stopping.

\section{Experimental Setup}

\subsection{Datasets}
The evaluation was performed on two main summarization datasets: CNN/DailyMail \citep{hermann2015teaching} and New York Times (NYT) \citep{sandhaus2008new}. 
CNN/DM contains news articles and associated highlights, i.e., a few bullet points giving a brief overview of the article. We used the standard splits of 287k, 13k, and 11k for training, validation, and testing, respectively. We did not anonymize entities and followed \citet{see2017get} to pre-process the first sentences of CNN. For NYT, we used a provided dataset used in \citet{liu2019text}, which consists of 38264, 4002, 3421 training, validation, and test set instances, respectively. The instances are news articles accompanied by short human-written summaries, where summaries shorter than 50 words were removed.

\begin{table}[t]
    \centering
    \begin{tabular}{c c c c}
    \thickhline
         Year &  Count & Avg. \# words & Avg. \# sents  \\ \thickhline 
         2016 & 12799 & 34.65 & 2.46 \\ 
         2017 & 11292 & 32.49 & 2.34 \\ \thickhline 
    \end{tabular}
    \caption{CNN summary statistics for more recent years.}
    \label{table:cnn-stats}
    \NoGap
\end{table}

The original CNN/DM dataset contains news from 2007 to 2015. To test whether transduction is beneficial for more recent news, we obtained the newer snapshots of CNN, namely for 2016 and 2017. We downloaded CNN articles published in 2016 and 2017 using NewsPlease,\footnote{\url{https://github.com/fhamborg/news-please}} extracted raw contents, and retained those having a story highlight as a summary in the beginning of the article. The statistics are shown in Table~\ref{table:cnn-stats}. These sets were used for transduction only. 

Finally, we truncated input documents to 1000 subwords \footnote{The maximum number of subwords includes sentence separator special tokens.} by preserving complete sentences. To monitor overfitting, we used 1000, 500, and 100 validation instances for transduction on CNN/DM, NYT, and CNN 2016/2017, respectively. In all experiments, we used ROUGE-L for the stopping criterion. For evaluation, we used the standard ROUGE package~\citep{lin2004rouge} and report F1 scores.

\subsection{Model Details}
For pre-initialization, we used the large pre-trained BART model \citep{lewis2019bart} available with FairSEQ. We also compare against BART that was fine-tuned on abstractive references (\textsc{BART + ft}). Further, we used a subword tokenizer with the maximum of 50k subwords. The model had 12 layers both in the encoder and decoder and a hidden size of 1024. In total, it consisted of 400M parameters. During fine-tuning and transduction, the architecture remained unchanged.

During joint fine-tuning ($\textsc{\ours{}}^-$), presented in Sec.~\ref{sec:joint_fine_tuning}, we masked 25\% of words in input articles, and set  $\alpha=0.1$ to produce on average 10\% of extractive instances at each epoch. In transduction, we masked 10\% of input words, and sampled 14k and 5k training instances at each epoch for CNN/DM and NYT, respectively. Here, $\alpha$ was set to 0.1. In all experiments, we used Adam \citep{kingma2014adam} for weight updates, and beam search for summary generation with 3-gram blocking \citep{paulus2017deep}. All experiments were performed on 8-GPU p3.8xlarge Amazon instance. For CNN/DM, we performed joint fine-tuning for 6 epochs and transduction for 3 epochs. For NYT, 9 epochs of joint fine-tuning and 3 epochs for transduction. 

\subsection{Extractive Summarizer}

To obtain extractive references for transduction (\textsc{\extref{}}), we used the BART’s fine-tuned encoder, and an additional transformer encoder \citep{vaswani2017attention} to contextualize sentence representations. For CNN/DM, we set the number of layers to 3, and attention heads to 16. For NYT, we set it to 2 layers with attention heads number to 8. To produce binary scores, we used a linear transformation that is followed by the sigmoid function. 

To select salient sentences from the training set input articles, we used a greedy heuristic (\textsc{Oracle}) that maximizes ROUGE scores between the salient sentences and the gold summary as in \citet{nallapati2016summarunner, liu2019text}. We selected up to 3 sentences per input article.
In inference, we ranked candidate sentences by scores and selected top-3 sentences. Also, we applied N-gram blocking during selection to avoid repetitive content as in \citet{liu2019text}. Given a current extractive summary $s$ and candidate sentence $c$, we skip $c$ if there exists a trigram overlap between $c$ and $s$. 

\subsection{Human Evaluation}
\label{sec:human_evaluation}
For human evaluation experiments, we randomly sampled 300 articles from CNN/DM test set. Further, we generated and compared summaries from \textsc{BART + ft} and \textsc{\ours{}}. We used Amazon Mechanical Turk (AMT) and ensured that only high-quality workers could participate. We asked workers to pass a custom qualification test, which only 14.6\% of those who took it passed. For further details, see Appendix~\ref{app:human_eval_crit}. Finally, we requested 3 annotators per HIT and used MACE~\citep{Hovy2013} to estimate annotator competences and recover the most likely answer per HIT accordingly. 

\begin{table*}[h!]
\centering
\begin{tabular}{ l c c c c c c }     
    \multicolumn{1}{c}{} & \multicolumn{3}{c}{\textbf{CNN/DailyMail}}  & \multicolumn{3}{c}{\textbf{New York Times}} \\
    \thickhline
      & R1 & R2 & RL & R1 & R2 & RL\\ \thickhline
    \textsc{Oracle} & 55.21 & 32.86 & 51.36 & 61.70 & 42.23 & 58.34 \\
    \textsc{Lead-3} & 40.42 & 17.62 & 36.67 & 38.28 & 19.75 & 34.96 \\ \thickhline
    \multicolumn{7}{c}{Extractive / Compressive} \\ \thickhline 
    \textsc{SummaRuNNer} \citep{nallapati2016summarunner} & 39.60 & 16.20 & 35.30 & - & - & -\\
    \textsc{Refresh} \citep{narayan2018ranking} & 40.00 & 18.20 & 36.60 & - & - & - \\
    \textsc{Sumo} \citep{liu2019single} & 41.00 & 18.40 & 37.20 & 42.30 & 22.70 & 38.60 \\
    \textsc{JETS} \citep{xu-durrett-2019-neural} & 41.70 & 18.50 & 37.90 & - & - & - \\
    \textsc{BertSumExt} \citep{liu2019text} & 43.25 & 20.24 & 39.63 & 46.66 & 26.35 & 42.62 \\
    \textsc{MatchSum} \citep{zhong-etal-2020-extractive} & 44.41 & 20.86 & 40.55 & - & - & - \\ \thickhline
    \multicolumn{7}{c}{Abstractive} \\ \thickhline 
    \textsc{PTGEN+COV} \citep{see2017get} & 39.53 & 17.28 & 36.38 & 43.71 & 26.40 & -\\
    \textsc{BottomUP} \citep{gehrmann-etal-2018-bottom} & 41.22 & 18.68 & 38.34 & - & - & -\\
    \textsc{DRM} \citep{paulus2017deep} & - & - & - & 42.94 & 26.02 & - \\
    \textsc{BertSumExtAbs} \citep{liu2019text} & 42.13 & 19.60 & 39.18 & 49.02 & 31.02 & 45.55 \\ 
    \textsc{PEGASUS} \citep{zhang2019pegasus} & 44.17 & 21.47 & 41.11 & - & - & -\\
    \textsc{BART + ft} (reported) \citep{lewis2019bart} & 44.16 & 21.28 & 40.90 & - & - & - \\
    \textsc{BART + ft} (ours)\tablefootnote{We used  shorter input with only complete sentences that we believe resulted in a slightly worse performance.} & 44.01 & 21.13 & 40.81 & 52.97 & 35.19 & 49.32 \\\thickhline
    
    \multicolumn{7}{c}{Ours} \\ \thickhline 
    \textsc{\ours{}}$^-$ & 44.59 & 21.58 & 41.50 & 53.55 & 35.54 & 49.81 \\
    \textsc{\ours{}} & \textbf{44.96}  & \textbf{21.89} & \textbf{41.86} & \textbf{53.72} & \textbf{35.72} & \textbf{50.06} \\
    \hline
    \textsc{\extref{}} & 43.93 & 21.12 & 40.20 & 47.49 & 27.57 & 43.88 \\
  \thickhline
    \end{tabular}
    \caption{ROUGE F1 scores on the standard CNN/DM and New York Times test sets.}
    \label{table:main-rouge-eval}
\end{table*}

\section{Evaluation Results}
\label{sec:eval}

\subsection{Automatic Evaluation}
\label{sec:automatic_eval}

\paragraph{Standard Datasets}
We report automatic evaluation based on ROUGE F1 on the CNN/DM and NYT test sets, the results are shown in Table~\ref{table:main-rouge-eval}.

First of all, we observed that joint fine-tuning (\textsc{\ours{}}$^-$) substantially outperforms the standard fine-tuning (\textsc{BART + ft}) despite their identical architectures. Second, we observed that transduction further improves the performance of the jointly fine-tuned model on both datasets (\textsc{\ours{}}). We also performed an independent-samples t-test to compare \textsc{\ours{}} to \textsc{BART+ft}. It indicates that all results are statistically significant under $p<0.05$ except ROUGE-2 on NYT.

In a qualitative analysis, we observed that \textsc{\ours{}} generates more coherent summaries with richer contexts and details. This is important for practical applications -- the user should get a full picture from the summary about a lengthy news article. See example summaries in Appendix.


\paragraph{Recent News}

It is common to assume that training and test sets share a common distribution~\citep{quadrianto2009distribution, kann2018neural}. Intuitively, this means that the training set well covers the content of the test set. In practice, however, future news often contain novel entities, people, and topics. This can make it difficult to generate high quality summaries for recent news when the summarizer was trained on dated ones. We explore this scenario and demonstrate the effectiveness of transduction. Specifically, we transduct models on CNN news from 2016 and 2017 while they were originally fine-tuned on CNN/DM news from 2007 to only 2015. The results are shown in Table~\ref{table:cnn-new-snapshots}.


First of all, we observed that joint fine-tuning is superior to the standard one on both datasets. Second, even though extractive noisy references (\textsc{\extref{}}) have low ROUGE scores, we further improve the results by performing transduction (\textsc{\ours{}}). 
Recall from Sec.~\ref{sec:joint_fine_tuning} that from the training set we select salient sentences from input articles by matching them to gold summaries. We experimented with the same approach on the test set to get higher quality extractive references for transduction (\textsc{Oracle}). As indicated by \textsc{\ours{}} w/ \textsc{Oracle}, when better extractive references are provided, it results in additional improvements. These results confirm that our approach is beneficial for settings where training set and test set distributions are different. 


\begin{table*}[h!]
\centering
    \begin{tabular}{  l  c  c  c  c  c  c }
    \multicolumn{1}{c}{} & \multicolumn{3}{c}{\textbf{CNN 2016}} & \multicolumn{3}{c}{\textbf{CNN 2017}} \\ \thickhline
     & R1 & R2 & RL & R1 & R2 & RL \\ \thickhline
    \textsc{Oracle} & 53.05 & 36.87 & 49.89 & 52.58 & 36.97 & 49.57 \\
    \textsc{Lead-3} & 31.87 & 16.62 & 29.06 & 28.82 & 14.32 & 26.21 \\ \thickhline
    \textsc{BERTSumExtAbs}  & 33.17 & 14.43 & 30.56 & 30.44 & 12.51 & 27.98 \\
    \textsc{BART + ft} & 34.93 & 15.83 & 32.14 & 32.62 & 14.27 & 29.98 \\
    \thickhline
    \textsc{\ours{}}$^-$ & 35.40 & 15.92 & 32.62 & 32.92 & 14.21 & 30.24 \\
    \textsc{\ours{}} & \textbf{35.58} & \textbf{16.32} & \textbf{32.78} & \textbf{33.07} & \textbf{14.63} & \textbf{30.45} \\
    \textsc{\ours{}} w/ \textsc{Oracle} & 36.10 & 16.72 & 33.27 & 33.37 & 15.01 & 30.71 \\
    \hline
    \textsc{\extref{}} & 32.14 & 15.37 & 29.17 & 29.19 & 13.30 & 26.43 \\ \thickhline
    \end{tabular}
    \caption{ROUGE F1 scores on more recent CNN test sets. In \textsc{\ours{}} \textbackslash w \textsc{Oracle} we used \textsc{Oracle} extractive references transduction.}
    \label{table:cnn-new-snapshots}
    \NoGap{}
\end{table*}

\subsection{Human Evaluation}
\label{sec:human_eval}

In manual investigations, we observed that \textsc{\ours{}} summaries are more coherent thus easier to read and better convey a story. In order to validate this observation, we performed a human evaluation study. Additionally, in another human evaluation, we focused on the factual consistency of summaries. Generated summaries should not introduce novel information not present in the input article as this can lead to user aversion. In turn, this is an open problem in summarization~\citep{kryscinski2019evaluating, brazinskas2020few, maynez2020faithfulness}.

\paragraph{Coherence} In evaluation, generated summaries were presented in a random order, as well as the input article and reference summary for context. For each HIT, we asked the 3 annotators which of the two generated summaries, if any, was more coherent. We gave the following definition: \textit{"The more coherent summary has better structure and flow, is easier to follow. The facts are presented in more logical order."}
The \textsc{\ours{}} model was preferred 110 times (22.0\%), while \textsc{BART + ft}  was preferred 89 times (17.8\%). In 101 cases (20.2\%), the annotators indicated that none of the two summaries was preferable. We conclude that the \textsc{\ours{}} summaries were significantly more coherent than the \textsc{BART + ft} summaries (p $<$ 0.05 using a one-sided z-test). 

We observed that CNN/DM articles tend to be more coherent than the associated bullet point summaries. Further, we observed that salient sentences we used for learning (\textsc{ExtRef}) tend to be among lead 5 (61.1\%) with a very small gap between them (0.529 sentences on average). Therefore, we hypothesize that the model learns from consecutive sentences more natural text structures emanating in summaries.

\paragraph{Factual Consistency} For evaluating factual consistency, each HIT presented one input article and one generated summary from \textsc{BART + ft} or \textsc{\ours{}}. To simplify the task, we focused the workers' attention on a single highlighted sentence per summary, which we picked at random, and asked if that sentence, as shown in the context of the full summary, is factually consistent with the article. We gave detailed guidelines and examples for factual errors, see Appendix~\ref{app:human_eval_crit}. Effectively, this setup measured how likely a randomly chosen summary sentence is factually consistent with the summarized article. We found that 263 of the 300 \textsc{BART + ft} summary sentences (87.7\%) were judged factual, compared to 254 for the 300 \textsc{\ours{}} summaries (84.7\%). This is a small difference that we found not statistically significant (p $<$ 0.05 using a one-sided z-test).

\section{Analysis}

\subsection{Transduction of Fine-tuned BART}

\begin{table}
\centering
    \begin{tabular}{ l  c  c  c }  \thickhline 
    & R1 & R2 & RL \\ \thickhline
    \textsc{BART + ft} & 44.01 & 21.13 & 40.81 \\
    \textsc{BART + ft + tr} & 44.83 & 21.79 & 41.69\\
    \thickhline
    \textsc{\ours{}}$^-$ & 44.59 & 21.58 & 41.50 \\
    \textsc{\ours{}} & \textbf{44.96}  & \textbf{21.89} & \textbf{41.86} \\
    \hline
    \thickhline
    \end{tabular}
    \caption{Comparison between transduction of the BART model that was fine-tuned using our and the default method on the CNN/DM test set.}
    \label{table:bart-ft-trans}
    \NoGap{}
\end{table}

We further explored how beneficial transduction is when applied to the standard model already fine-tuned on abstractive summaries -- (\textsc{BART + ft}). The results on the standard CNN/DM dataset are presented in Table~\ref{table:bart-ft-trans}. They indicate that transduction is substantially beneficial and noticeably improves the results. We hypothesize that the model also benefits from the training set extractive instances that are jointly predicted. Nevertheless, it does not reach the results achieved by our full approach -- \textsc{\ours{}}.

\subsection{Ablation}
\label{sec:ablation}

\begin{figure*}[htb]

    \begin{subfigure}{0.5\textwidth}
        \centering
        \includegraphics[width=1.\linewidth]{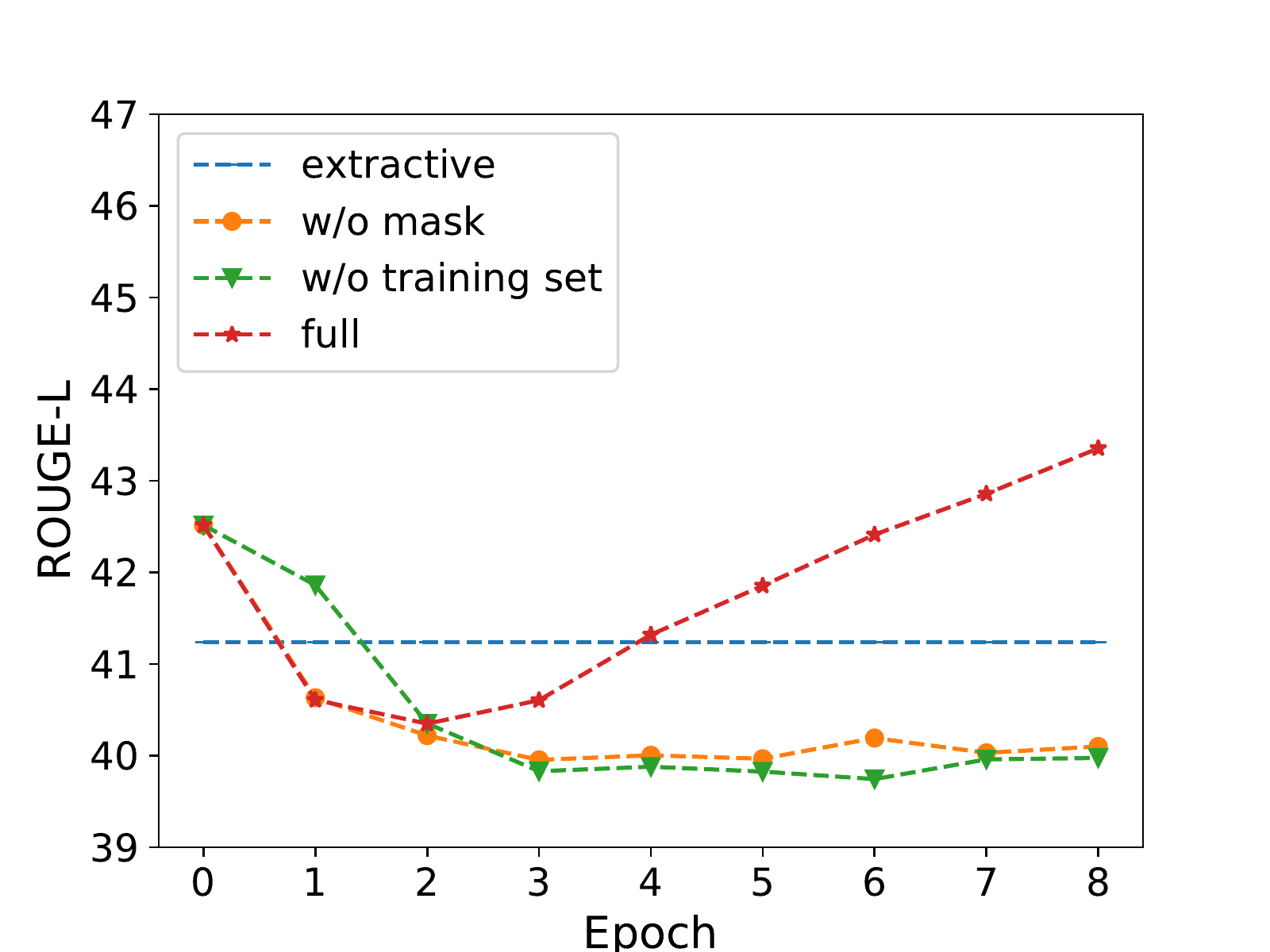}
        \caption{\textsc{BART + ft}}
        \label{fig:ft-tr-ablation}
    \end{subfigure}%
    \begin{subfigure}{0.5\textwidth}
        \centering
        \includegraphics[width=1.\textwidth]{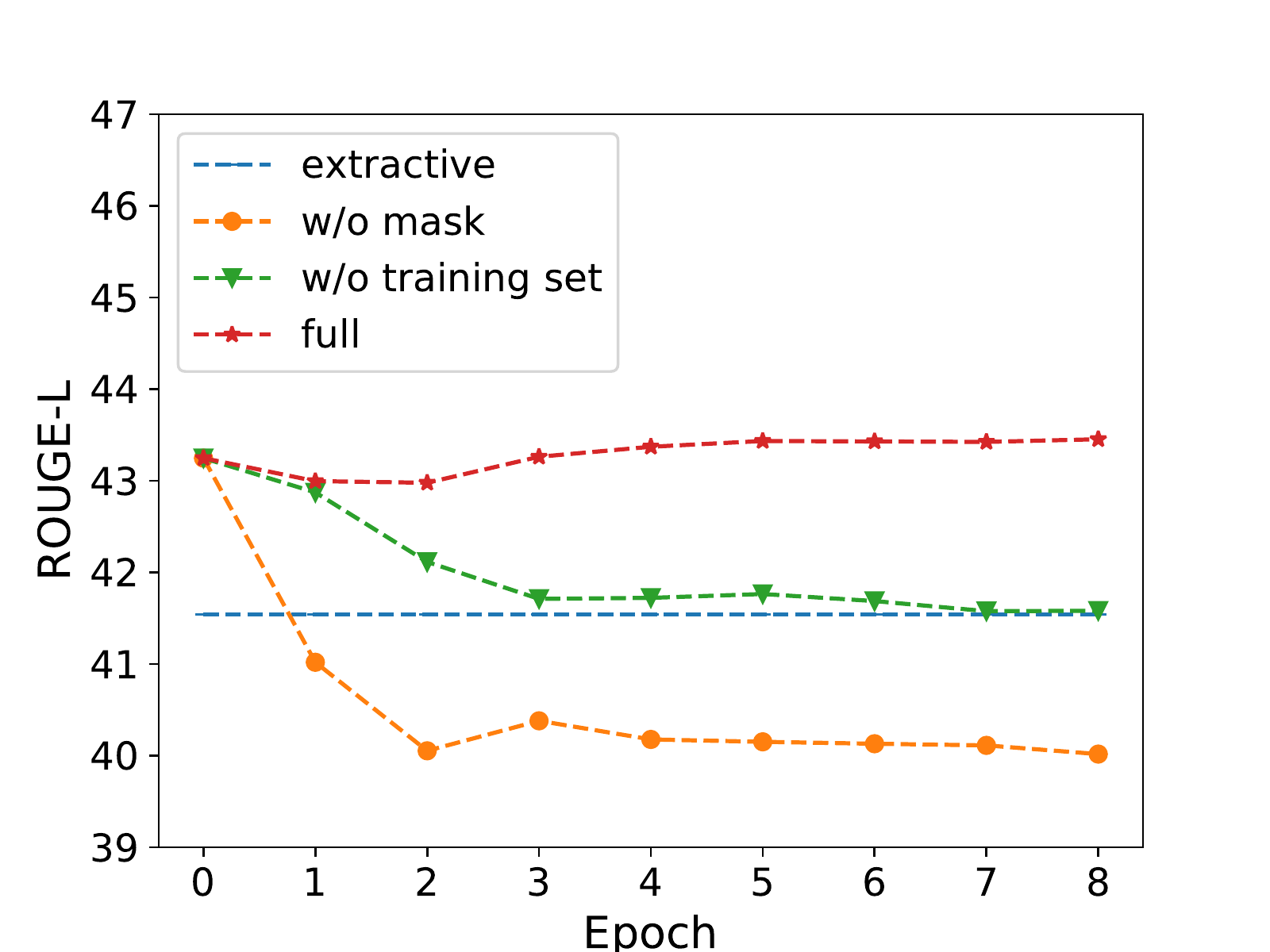}
        \caption{\textsc{\ours{}}}
        \label{fig:mft-tr-ablation}
    \end{subfigure}
    \caption{Ablation during the transduction phase. ROUGE-L scores on the 1k subset of the standard CNN/DM validation set; \textit{extractive} indicates the extractive references used for transduction.}
    \label{fig:ablation}
\end{figure*}

To gain insights into the inner workings of transduction, we performed an ablation study by removing various components during the transduction phase. We plot the ROUGE-L scores on the validation subset in Fig.~\ref{fig:ablation}.  

First of all, we did not mask the input article and observed that both models degradate. We believe that the mask token is used as a mode indicator for the decoder. And without it, the decoder is unable to differentiate the two modes (extractive vs abstractive summary prediction). Second, we observed that the removal of the training set instances for regularization, as explained in Sec.~\ref{sec:trans-reg} has also a strong effect on the performance. Specifically, it makes \textsc{\ours{}}$^-$ converge to the same ROUGE score as the extractive references used for transduction and \textsc{BART + ft} degradate. Finally, without ablations, we observed two different learning dynamics. Specifically, \textsc{BART + ft} initially decreases in the ROUGE score for 2 epochs, and then slowly starts to improve by surpassing the baseline extractive references at epoch 4. We hypothesize that it is caused by unfamiliarity with predicting extractive summaries. In turn, \textsc{\ours{}} experiences only a minor decrease in the beginning, possibly due to the lower quality of extractive references of the test set tagged by the model in Sec.~\ref{sec:extr-summ}, and then it steadily improves.



\subsection{Novel N-grams}
\label{sec:novel-ngrams}

We also analyzed generated summaries in terms of the proportion of novel $n$-grams that appear in the produced summaries but not in the source texts.  The results are shown in Table \ref{table:novel-ngrams}. We observed that joint fine-tuning and transduction increase the proportion of novel $n$-grams, thus making summaries more abstractive. By comparing extractive and abstractive summaries, we noticed the selected sentences in extractive summaries often paraphrase sentences in the abstractive ones. We hypothesize that the exposure to the references with paraphrases allows the model to generate more variant summaries.  


\begin{table}
\centering
    \begin{tabular}{l c c c c}     \thickhline
     & N1 & N2 & N3\\ \thickhline
     Gold & 0.178 & 0.528 & 0.718 \\
     \thickhline
     \textsc{BART + ft} & 0.019 & 0.101 & 0.186  \\
     \textsc{BART + ft + tr} & 0.026 & 0.132 & 0.234 \\
     \hline
     \textsc{\ours{}}$^{-}$ & 0.028 & 0.135 & 0.238 \\
     \textsc{\ours{}} & \textbf{0.029} & \textbf{0.145} & \textbf{0.254} \\
\thickhline
    \end{tabular}
    \caption{The proportion of novel $n$-grams on the standard CNN/DM test set.}
    \label{table:novel-ngrams}
    \NoGap{}
\end{table}

\section{Related Work}
\label{sec:rel_work}

Single-document extractive and abstractive summarization is a well-established field with a large body of prior research~\citep{dasgupta-kumar-ravi-acl2013, rush2015neural, nallapati2016abstractive, tan-etal-2017-abstractive, see2017get, fabbri2020improving, laban2020summary}. 
 
The utilization of extractive summaries to improve abstractive summarization has also received some recent attention. Commonly, in a two-step procedure where salient fragments are first selected, and then paraphrased into abstractive summaries \citep{Chen2018FastAS, bae2019summary}. Alternatively, to alter attention weights \citep{hsu2018unified, gehrmann-etal-2018-bottom} to bias the model to rely more on salient input content. Finally, to perform pre-training on extractive references prior to abstractive summarization \citep{liu2019text}. In our case, we predict extractive references word-by-word by constructing a denoising objective that also masks input words. We use the same model without modifications, and predict extractive and abstractive references jointly.

Transductive learning has been applied to a number of language-related tasks, such as machine translation \citep{ueffing2007transductive}, paradigm completion \citep{kann2018neural}, syntactic and semantic analysis \citep{ouchi2019transductive}, and more recently to style transfer \citep{xiao2021transductive}. However, to the best of our knowledge, transductive learning has never been applied to summarization.

More recently, \textsc{PEGASUS} \citep{zhang2019pegasus} leveraged text fragments for pre-training. The text fragments are selected using heuristics, such as top-K sentences. Instead, we utilize a separate extractive model or gold summaries to select sentences that form extractive references. Another related approach that has been proven to be beneficial -- domain adaptation via task-specific pre-training~\citep{gururangan-etal-2020-dont}. However, they don't consider utilizing test set inputs for adaptation.  


\section{Conclusions}

In this work, we present the first application of \textit{transductive learning} to summarization. We propose learning from salient sentences extracted from the test set's input articles to better capture their specifics. We additionally propose a mechanism to regularize and validate the transductive model. The proposed method achieves state-of-the-art results in automatic evaluation on the CNN/DM and NYT datasets, and it generates more abstractive and coherent summaries. Finally, we demonstrate that transduction is useful when trained on dated news and transducted on more recent news.

\section{Future Work}
\label{sec:future_work}
Transductive learning in the context of summarization is an exciting research avenue. In this section, we propose a number of potential future directions. First, learning from single data points in the online fashion can be a promising direction. This, in turn, could call for the decoder's modularization that is less prone to overfitting. This could be achieved using more efficient fine-tuning methods, such as adapters~\citep{houlsby2019parameter} and continuous prefixes~\citep{li-liang-2021-prefix}.
Second, we hypothesize that content fidelity can be improved by learning from the test set inputs using specialized methods. For instance, one could learn to predict entities present in news articles to reduce hallucinations, similar to \citet{narayan2021planning}. Third, where training and test sets are in different domains, adaptation in the transduction phase can be fruitful, similar to \citet{ueffing2007transductive}. This follows the direction of domain adaptation via task-specific pre-training~\citep{gururangan-etal-2020-dont}.  


\section{Ethics Statement}
\label{sec:ethics}

\paragraph{Human Evaluation}
We used a publicly available service (Amazon Mechanical Turk) to hire voluntary participants, requesting native speakers of English. The participants were compensated above the minimum hourly wage in their self-identified countries of residence. 

\paragraph{Dataset}

The dataset was collected and used in accordance to non-commercial personal purpose permitted by the data provider.  



\bibliography{anthology,custom}
\bibliographystyle{acl_natbib}

\section{Appendices}
\label{sec:appendix}

\subsection{Details on the Mechanical Turk Setup}
\label{app:human_eval_crit}

\paragraph{Custom Qualification Test.} For all our evaluations on Mechanical Turk, we first created a custom qualification test that could be taken by any worker from a country whose main language is English, who has completed 100 or more HITs so far with an acceptance rate of 95\% or higher. The qualification test consisted of three questions from our factual consistency setup; two of which had to be answered correctly, along with an explanation text (5 words or more) to explain when "not factually consistent" was chosen. 53\% of workers who started the test provided answers to all three questions, and 27.6\% of these answered at least two correctly and provided a reasonable explanation text, i.e., only 14.6\% of the test takers were granted the qualification. The qualification enabled workers to work on our factual consistency HITs as well as our HITs judging summary coherence. 

\begin{figure}[t]
    \centering
    \includegraphics[width=\columnwidth]{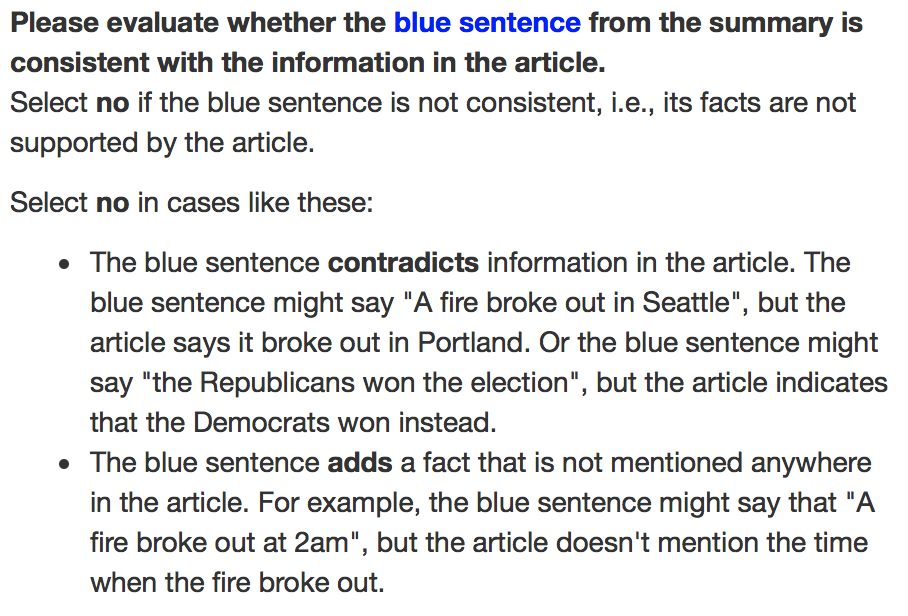}
    \caption{Instructions for evaluating if a summary sentence (highlighted in blue) was factually consistent with the source article.}
    \label{fig:factual_instructions}
\end{figure}

\paragraph{Payment and Instructions.} The coherence task took workers a median time of 125 seconds per HIT, for which we paid \$0.40 with a bonus pf \$0.20, amounting to an hourly rate of \$17. 
The factual consistency task took workers a median time of 30 seconds per summary; the payment was \$0.12 plus a bonus of \$0.05, amounting to an hourly rate of \$20. This task was relatively quick to do as a single summary sentence had to be judged; we also highlighted article sentences that are semantically similar to the highlighted summary sentence, in order to make the relevant information from the article more quickly accessible for fact checking.\footnote{We used the cosine distance of the universal sentence embeddings \cite{Cer2018} to measure semantic similarity.}
The factual consistency task contained instructions shown in Fig.~\ref{fig:factual_instructions}. The instructions for the coherence task are quoted in the main text above.

\paragraph{Excluding Spammers.} For both tasks, we ran code attempting to automatically detect potential spammers and label them for exclusion, in order to ensure high quality annotations. Anyone labeled for exclusion was disqualified for further HITs, their HIT answers were excluded from the results and HITs were extended to seek replacement answers. For the coherence task, any worker who spent less than 10 seconds per HIT was labeled for exclusion. For the factual consistency task, the minimum time per HIT required was 5 seconds; in addition; workers who wrote very short explanation texts for their "not factually consistent" answers (median length 3 words or less) were excluded. We also added 10 HITs with known factuality, and workers who answered 3 or more of them but with an accuracy less than 2/3 were excluded as well. Any worker who was not excluded according to the above criteria received the bonus.

\paragraph{Inter-Annotator Agreement and MACE} For the binary factual consistency evaluation, 521 of the 600 HITs (86.8\%) had a full agreement of all 3 workers; all other HITs had two agreements. For the coherence evaluation, in which 3 different answers were possible (first or second summary more coherent; or none), 258 of the 300 HITs (86.0\%) had an agreement of 2 or more workers per HIT. As noted in the main text above, we ran MACE \cite{Hovy2013} to further improve upon these raw answers by unsupervised estimation of worker trustworthiness and subsequent recovery of the most likely final answer per HIT.

\begin{table*}    \centering
 	\footnotesize 
    \begin{tabular}{  >{\centering\arraybackslash} m{2cm} m{12cm}}
    \thickhline 
    \textsc{Gold} & \vspace{0.5em} 
    College-bound basketball star asks girl with Down syndrome to high school prom . Pictures of the two during the "prom-posal" have gone viral .
 \vspace{0.5em} \\ \hline
    \textsc{BART + ft} & \vspace{0.5em} 
    Trey Moses asked Ellie Meredith to be his prom date. Ellie has Down syndrome. Trey is a star basketball player at Eastern High School in Louisville, Kentucky. Photos of the couple have gone viral on Twitter. "That's the kind of person Trey is," Ellie's mom says.
 \vspace{0.5em} \\ \hline
    \textsc{TrSum} & \vspace{0.5em} 
    Trey Moses, a star basketball player, asks Ellie Meredith, a freshman with Down syndrome, to prom . "She's great... she listens and she's easy to talk to," Trey says . Photos of the couple have gone viral on Twitter . Ellie's mom: "You just feel numb to those moments raising a special needs child"
    \vspace{0.5em} 
    \\ \thickhline 
    \textsc{Article} & \vspace{0.5em} 
    He's a blue chip college basketball recruit. She's a high school freshman with Down syndrome. At first glance Trey Moses and Ellie Meredith couldn't be more different. But all that changed Thursday when Trey asked Ellie to be his prom date. Trey -- a star on Eastern High School's basketball team in Louisville, Kentucky, who's headed to play college ball next year at Ball State -- was originally going to take his girlfriend to Eastern's prom. So why is he taking Ellie instead? "She's great... she listens and she's easy to talk to" he said. Trey made the prom-posal (yes, that's what they are calling invites to prom these days) in the gym during Ellie's P.E. class. Trina Helson, a teacher at Eastern, alerted the school's newspaper staff to the prom-posal and posted photos of Trey and Ellie on Twitter that have gone viral. She wasn't surpristed by Trey's actions. "That's the kind of person Trey is," she said. To help make sure she said yes, Trey entered the gym armed with flowers and a poster that read "Let's Party Like it's 1989," a reference to the latest album by Taylor Swift, Ellie's favorite singer. Trey also got the OK from Ellie's parents the night before via text. They were thrilled. "You just feel numb to those moments raising a special needs child,"  said Darla Meredith, Ellie's mom. "You first feel the need to protect and then to overprotect." Darla Meredith said Ellie has struggled with friendships since elementary school, but a special program at Eastern called Best Buddies had made things easier for her. She said Best Buddies cultivates friendships between students with and without developmental disabilities and prevents students like Ellie from feeling isolated and left out of social functions. "I guess around middle school is when kids started to care about what others thought," she said, but "this school, this year has been a relief." Trey's future coach at Ball State, James Whitford, said he felt great about the prom-posal, noting that Trey, whom he's known for a long time, often works with other kids . Trey's mother, Shelly Moses, was also proud of her son. "It's exciting to bring awareness to a good cause," she said. "Trey has worked pretty hard, and he's a good son." Both Trey and Ellie have a lot of planning to do. Trey is looking to take up special education as a college major, in addition to playing basketball in the fall. As for Ellie, she can't stop thinking about prom. "Ellie can't wait to go dress shopping" her mother said. "Because I've only told about a million people!" Ellie interjected.
    \vspace{0.5em} 
    \\ \thickhline    
    \end{tabular}
    \caption{Example summaries generated by \textsc{BART + ft} and \textsc{\ours{}}. We often observed that \textsc{\ours{}} summaries are more detailed, coherent, and with richer contexts. }
\end{table*}

\begin{table*}    \centering
 	\footnotesize 
    \begin{tabular}{  >{\centering\arraybackslash} m{2cm} m{12cm}}
    \thickhline 
    \textsc{Gold} & \vspace{0.5em} 
    "Furious 7" pays tribute to star Paul Walker, who died during filming . Vin Diesel: "This movie is more than a movie" "Furious 7" opens Friday .
 \vspace{0.5em} \\ \hline
    \textsc{BART + ft} & \vspace{0.5em} 
    Paul Walker died in a car crash in November 2013. The film "Furious 7" is out Friday. There have been multiple tributes to Walker leading up to the release. "You'll feel it when you see it," says co-star Vin Diesel. "It's bittersweet, but I think Paul would be proud"
 \vspace{0.5em} \\ \hline
    \textsc{TrSum} & \vspace{0.5em} 
    Paul Walker's death in November 2013 was especially eerie given his rise to fame in the "Fast and Furious" film franchise . The release of "Furious 7" on Friday offers the opportunity for fans to remember the man that so many have praised . "He was a person of humility, integrity, and compassion," said military veteran Kyle Upham .
    \vspace{0.5em} 
    \\ \thickhline 
    \textsc{Article} & \vspace{0.5em} 
Paul Walker is hardly the first actor to die during a production. But Walker's death in November 2013 at the age of 40 after a car crash was especially eerie given his rise to fame in the "Fast and Furious" film franchise. The release of "Furious 7" on Friday offers the opportunity for fans to remember -- and possibly grieve again -- the man that so many have praised as one of the nicest guys in Hollywood. "He was a person of humility, integrity, and compassion," military veteran Kyle Upham said in an email to CNN. Walker secretly paid for the engagement ring Upham shopped for with his bride. "We didn't know him personally but this was apparent in the short time we spent with him. I know that we will never forget him and he will always be someone very special to us," said Upham. The actor was on break from filming "Furious 7" at the time of the fiery accident, which also claimed the life of the car's driver, Roger Rodas. Producers said early on that they would not kill off Walker's character, Brian O'Connor, a former cop turned road racer. Instead, the script was rewritten and special effects were used to finish scenes, with Walker's brothers, Cody and Caleb, serving as body doubles. There are scenes that will resonate with the audience -- including the ending, in which the filmmakers figured out a touching way to pay tribute to Walker while "retiring" his character. At the premiere Wednesday night in Hollywood, Walker's co-star and close friend Vin Diesel gave a tearful speech before the screening, saying "This movie is more than a movie." "You'll feel it when you see it," Diesel said. "There's something emotional that happens to you, where you walk out of this movie and you appreciate everyone you love because you just never know when the last day is you're gonna see them." There have been multiple tributes to Walker leading up to the release. Diesel revealed in an interview with the "Today" show that he had named his newborn daughter after Walker. Social media has also been paying homage to the late actor. A week after Walker's death, about 5,000 people attended an outdoor memorial to him in Los Angeles. Most had never met him. Marcus Coleman told CNN he spent almost \$1,000 to truck in a banner from Bakersfield for people to sign at the memorial. "It's like losing a friend or a really close family member ... even though he is an actor and we never really met face to face," Coleman said. "Sitting there, bringing his movies into your house or watching on TV, it's like getting to know somebody. It really, really hurts." Walker's younger brother Cody told People magazine that he was initially nervous about how "Furious 7" would turn out, but he is happy with the film. "It's bittersweet, but I think Paul would be proud," he said. CNN's Paul Vercammen contributed to this report.

    \vspace{0.5em} 
    \\ \thickhline    
    \end{tabular}
    \caption{Example summaries generated by the standard model -- \textsc{BART + ft} -- and our proposed one -- \textsc{\ours{}}. In this particular instance, our model generates a more coherent summaries with more details.}
\end{table*}

\end{document}